\documentclass[10pt,twocolumn,letterpaper]{article}

\usepackage{iccv}
\usepackage{times}
\usepackage{epsfig}
\usepackage{graphicx}
\usepackage{amsmath}
\usepackage{amssymb}
\usepackage{bm}
\usepackage{multirow}

\usepackage[pagebackref=true,breaklinks=true,letterpaper=true,colorlinks,bookmarks=false]{hyperref}

\iccvfinalcopy 

\def\iccvPaperID{901} 

\ificcvfinal\fi
\begin{document}

\title{Ordinal Distribution Regression for Gait-based Age Estimation}

\author{Haiping Zhu\textsuperscript{1}, Yuheng Zhang\textsuperscript{1}, Guohao Li\textsuperscript{1}, Junping Zhang\textsuperscript{1}, and Hongming Shan\textsuperscript{1}\\
\textsuperscript{1}Shanghai Key Lab of Intelligent Information Processing, \\
School of Computer Science, Fudan University, \\
Shanghai, China, 200433 \\
\{hpzhu14, yuhengzhang16, ghli17, jpzhang, hmshan\}@fudan.edu.cn}

\maketitle

\begin{abstract}
Computer vision researchers prefer to estimate age from face images because facial features provide useful information. However, estimating age from face images becomes challenging when people are distant from the camera or occluded. A person’s gait is a unique biometric feature that can be perceived efficiently even at a distance. Thus, gait can be used to predict age when face images are not available. However, existing gait-based classification or regression methods ignore the ordinal relationship of different ages, which is an important clue for age estimation. This paper proposes an ordinal distribution regression with a global and local convolutional neural network for gait-based age estimation. Specifically, we decompose gait-based age regression into a series of binary classifications to incorporate the ordinal age information. Then, an ordinal distribution loss is proposed to consider the inner relationships among these classifications by penalizing the distribution discrepancy between the estimated value and the ground truth. In addition, our neural network comprises a global and three local sub-networks, and thus, is capable of learning the global structure and local details from the head, body, and feet. Experimental results indicate that the proposed approach outperforms state-of-the-art gait-based age estimation methods on the OULP-Age dataset.
\end{abstract}

\section{Introduction}
Human age estimation is an active research topic because age estimation plays an important role in many applications, such as video surveillance, social networking, and human–computer interaction. Existing age estimation methods are primarily based on facial images~\cite{chang2011ordinal,chen2017using,niu2016ordinal,pan2018mean,shen2017deep}, which are very informative and easy to be estimated. However, the performance of face-based age estimation approaches will be compromised when the face is occluded, for example, by sunglasses or makeup. In addition, face-based age estimation becomes challenging if a person is far away from the camcorders, which often occurs in many video surveillance systems located at crossroads, airports, and railway stations. As a unique biometric feature that can be perceived efficiently at a distance, using a person’s gait can be an alternative way to predict age when a facial image is not sufficiently informative or not available. Gait-based estimation has a psychological foundation that cannot be easily faked~\cite{han2006individual}. For example, an old person might walk slowly and hobble, whereas a young person might walk briskly.

\begin{figure}[!t]
\centering
\includegraphics[width=1\linewidth,height=0.4\linewidth, clip=true, trim=60 160 80 30]{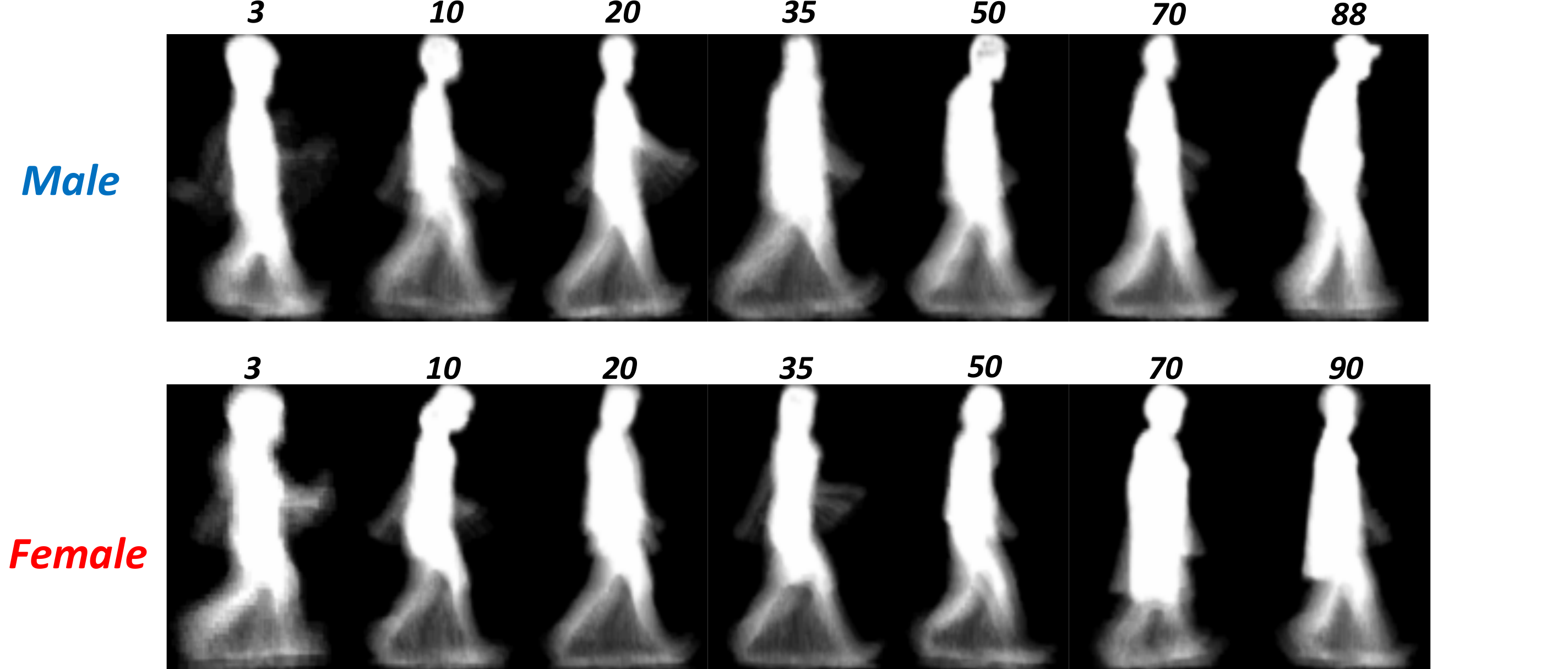}
\caption{GEIs of subjects of different age and gender in the OULP-Age dataset. The number above each GEI indicates the corresponding age of the subject.}
\label{gait_examples}
\end{figure}

\begin{figure}[!t]
\centering
\includegraphics[width=1\linewidth,height=0.58\linewidth, clip=true, trim=20 160 50 40]{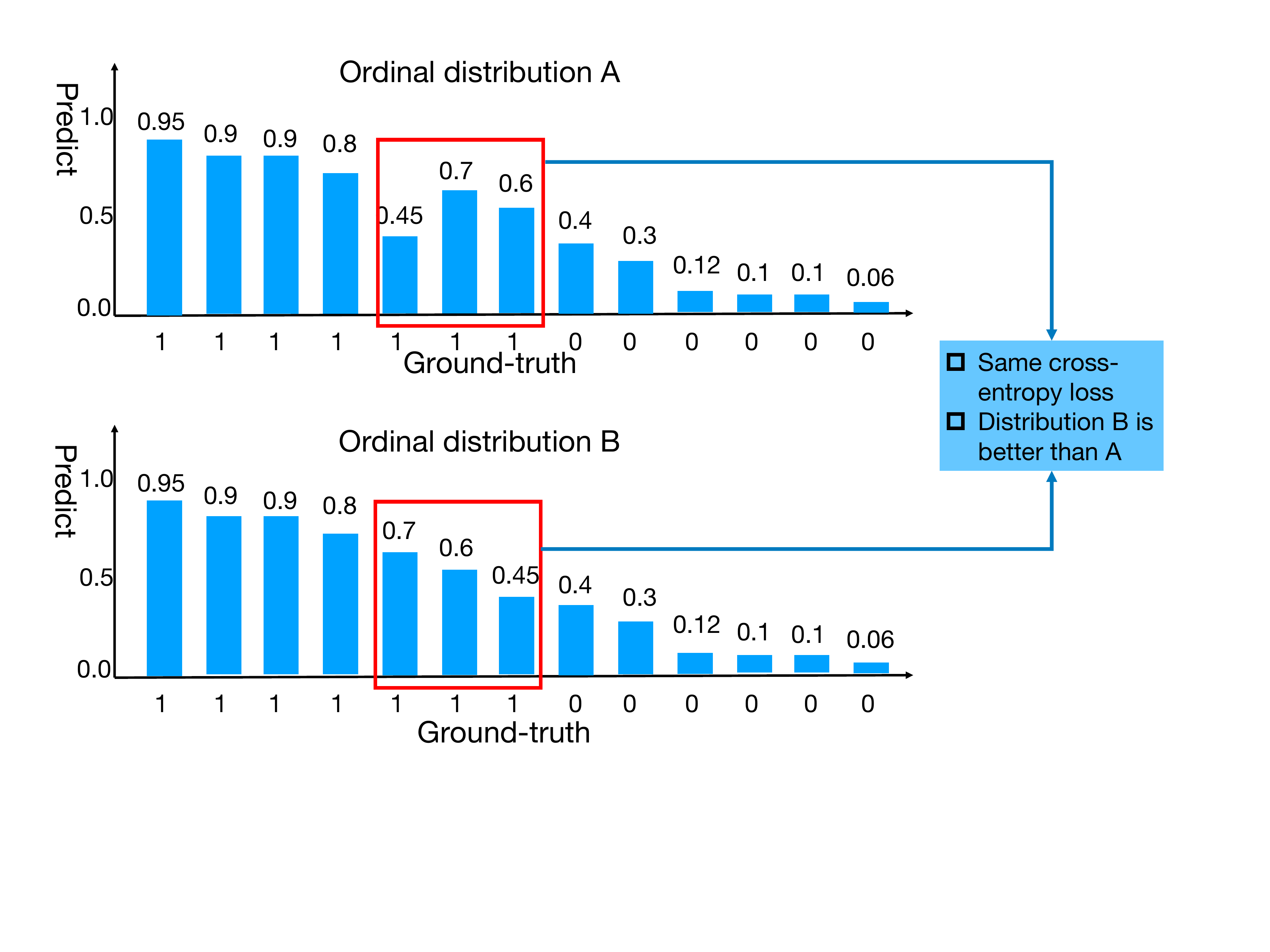}
\caption{Predictive probability of the $k$-th classifier should not be greater than that of the ($k-1$)-th classifier on an ordinal distribution. Both A and B have the same cross-entropy loss; however, B is preferable to A on an ordinal distribution.}
\label{toy_example}
\end{figure}

In the field of gait-based age estimation, gait energy image (GEI)~\cite{lu2010gait,makihara2011gait}, which compresses one or more gait sequences into a single image (as shown in Fig.~\ref{gait_examples}), is one of the most widely used gait templates due to its simplicity and effectiveness. Some studies have applied age manifold learning techniques on GEIs to learn a low-dimensional representation that captures the intrinsic data distribution and geometric structure~\cite{lu2010ordinary,lu2013ordinary}. Existing gait-based age estimation approaches can be roughly categorized as classification-based ~\cite{lu2010gait} and regression-based methods~\cite{lu2013ordinary,makihara2011gait}. However, neither of them considers the ordinal relationship between age labels, which is an important clue for age estimation. Therefore, to address this issue, ranking-based methods ~\cite{chen2017using,frank2001simple,li2007ordinal,niu2016ordinal} have been proposed. Typically, these methods decompose the ordinal regression into a series of binary classifications and utilize the cross-entropy loss to optimize binary classifications. However, the cross-entropy loss treats these classifications independently, ignoring the inner relationships among them. For these ordered binary classifications, the expected inner relationship is that the predictive probability of the $k$-th classifier should not be greater than the probability of the ($k-1$)-th classifier, as explained in Fig.~\ref{toy_example}.

This paper proposes an ordinal distribution regression with a global and local convolutional neural network, which we refer to as ODR-GLCNN, for gait-based age estimation. Similar to ranking-based methods for facial-based age estimation, we consider gait-based age estimation as an ordinal regression and decompose the ordinal regression problem into a series of binary classification sub-problems. Note that the primary issue with existing ranking-based methods is that they solve these binary sub-problems \emph{independently}. Consequently, such methods neglect the inner relationship to some extent and do not make good use of the correlation between these sub-binary tasks. To address this limitation, we propose an ordinal distribution loss (ODL) to penalize the distribution difference between the estimated and ground-truth age. Similar to Hou \emph{et al.}'s work~\cite{hou2016squared}, this proposed loss also utilizes the squared Earth mover's distance (EMD) to penalize the distribution discrepancy. Unlike Hou \emph{et al.}~\cite{hou2016squared}, which utilizes the squared EMD loss on the classification task, we use the loss in an ordinal regression task, {\emph e.g.}, gait-based age estimation. To the best of our knowledge, this is the first time that the distribution discrepancy on ordinal regression through squared EMD loss has been considered. In addition, we propose a unique network, consisting of a global and three local sub-networks, to obtain a global structure and local structures from the head, body, and feet. Experimental results on the OULP-Age dataset~\cite{xu2017isir} and the MORPH Album II dataset~\cite{ricanek2006morph} demonstrate that the proposed approach outperforms state-of-the-art methods on both gait-based and face-based age estimation.

The contributions of this paper are: 1) A deep ordinal distribution regression for gait-based age estimation is proposed, which achieves state-of-the-art predictive performance on the OULP-Age dataset; 2) A novel network, named as GL-CNN, comprising a global network and three local sub-networks is proposed to learn more representative features from the gait globally and locally; 3) An ordinal distribution loss (ODL) is proposed to consider the inner relationships among a series of binary sub-problems.

\section{Related work}
In this section, we provide a brief survey of face- and gait-based age estimation studies as well as ordinal regression studies.

\textbf{Face-based age estimation:}
Existing face-based age estimation approaches can be categorized as classification-, regression-, and ranking-based methods. Classification-based methods are often used to roughly estimate the age group of a subject in a face image~\cite{lanitis2004comparing,yang2007demographic}. Different ages or age groups are treated as independent classes. However, these methods do not adequately consider the cost difference of subjects belonging to different age groups. Regression-based methods provide a more accurate age assessment from a facial image ~\cite{fu2008human,xiao2009metric}. Typically, regression-based methods employ Euclidean loss ($\ell_2$ loss) to penalize the difference between the estimated age and the ground-truth age. Recently, ranking-based or ordinal methods have been proposed for facial age estimation~\cite{chang2015learning,chang2011ordinal,niu2016ordinal,chen2017using}. These approaches consider age as an ordered label and use multiple binary classifiers to determine the rank of a specific age. Unlike the $\ell_2$ loss, which ignores the ordinal information, ranking-based methods can explicitly model the ordinal relationships among face images sampled from different ages.

\textbf{Gait-based age estimation:}
The earliest gait-based age estimation study was that of ~\cite{makihara2011gait}, where a Gaussian process regression (GPR)~\cite{rasmussen2004gaussian} was introduced to predict age from human gait. Then, GPR was refined with an active set method~\cite{wada2013gaussian} to reduce the computational time required for online age estimation~\cite{makihara2016gait}. Lu and Tan proposed a multi-label guided (MLG) subspace  to better characterize the feature space by correlating the age and gender information of subjects~\cite{lu2010gait}. They further proposed an ordinary preserving manifold learning approach to seek a low-dimensional discriminative subspace for age estimation~\cite{lu2013ordinary}. Considering the age variations within different groups, such as children, adults, and the elderly, Li \emph{et al.} proposed an age group-dependent manifold method~\cite{li2018gait}. In that method, after an age group classifier was trained, a kernel support vector machine (SVM) regression was added to provide accurate assessment in each age group. This method achieved state-of-the-art gait-based age estimation performance.

\begin{figure*}[!ht]
\centering
\includegraphics[width=0.88\linewidth,height=0.5\linewidth, clip=true, trim=10 190 50 0]{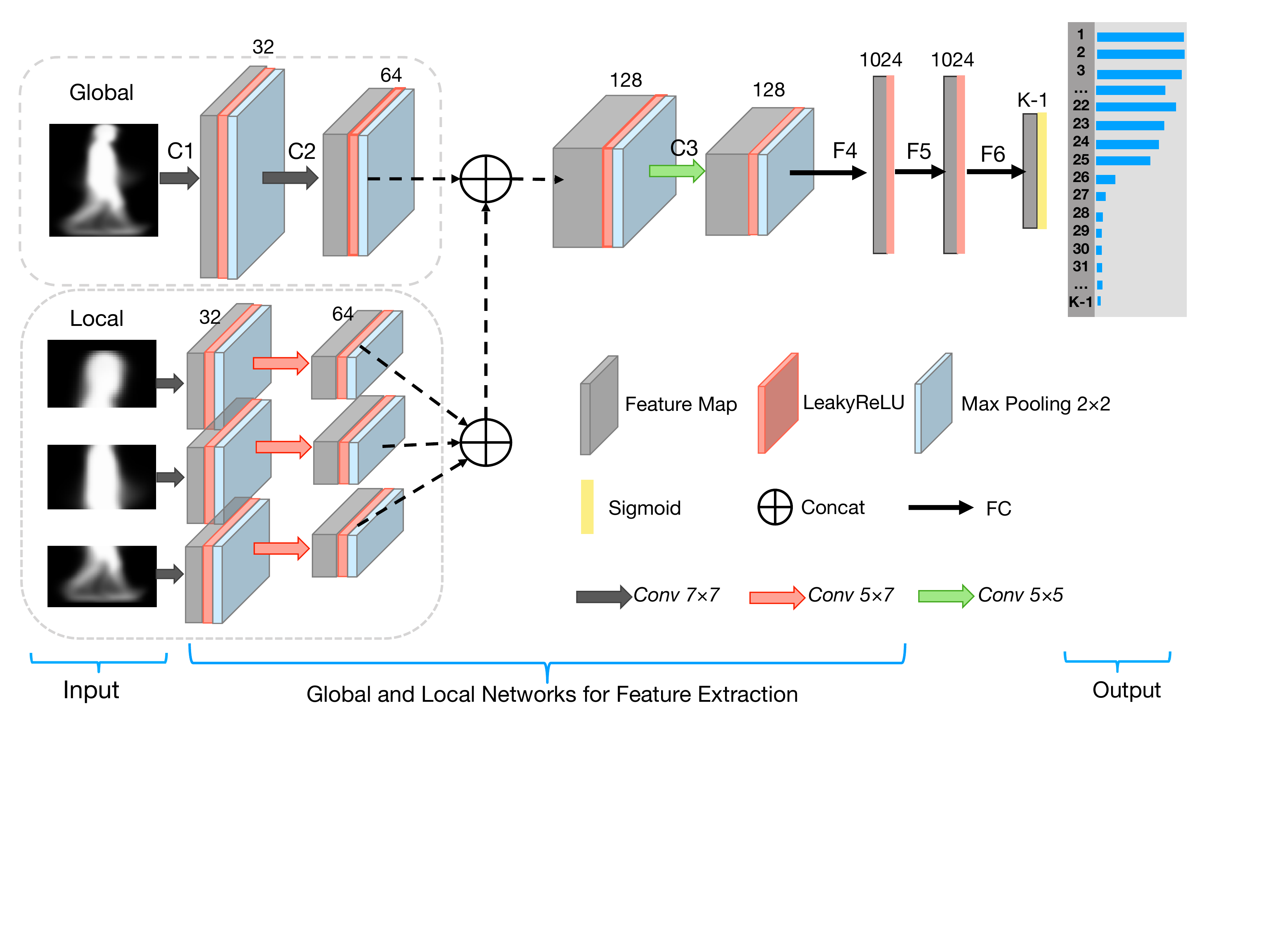}
\caption{Structure of the proposed ODR-GLCNN. The output layer contains $K-1$ binary classifications incorporating the ordinal information into the current end-to-end learning process.}
\label{structure}
\end{figure*}

\textbf{Ordinal regression:}
Most ordinal algorithms can be considered as refined versions of classification algorithms with ordinal constraints~\cite{crammer2002pranking,herbrich1999support,shashua2003ranking}. For example, Herbrich \emph{et al.} utilized an SVM for ordinal regression~\cite{herbrich1999support}. Then, Shashua and Levin refined the SVM to handle multiple thresholds~\cite{shashua2003ranking}. Crammer and Singer proposed the perceptron ranking algorithm to generalize an online perceptron algorithm with multiple thresholds for ordinal regression~\cite{crammer2002pranking}. Another way to utilize classification algorithms directly is to transfer ordinal regression into a series of simpler binary classifications~\cite{frank2001simple,li2007ordinal}. Specifically, Frank and Hall utilized a decision tree for binary classifications for ordinal regression~\cite{frank2001simple}. Li and Lin learned ordinal regression using a set of classifiers, and then employed an SVM for the final classification~\cite{li2007ordinal}. Recently, Niu \emph{et al.} introduced a CNN with multiple binary outputs to solve the ordinal regression for age estimation~\cite{niu2016ordinal}. Ordinal regression was also used in~\cite{chen2017using} by learning multiple binary CNNs and aggregating the final outputs. However, these ordinal regression methods solve each binary sub-problem independently and do not consider the underlying relationships among these binary sub-problems. Thus, this paper proposes a distribution loss to utilize such a relationship to improve the accuracy of age estimation.

\section{Proposed method}
In this section, we describe the proposed ordinal regression for gait-based age estimation, the network comprising a single global and three local sub-networks, and a unique distribution loss in more detail.

\subsection{Ordinal regression}
We treat gait-based age estimation as an ordinal regression in order to utilize the ordinal relationship of age labels. Let $\bm{x}_i\in \mathcal{X}$ denote the $i$-th input GEI sample. The corresponding age is $y_i\in\mathcal{Y}=\{r_1, r_2, \dots, r_K\}$ with ordered ranks $r_K \succeq r_{K-1} \succeq \dots \succeq r_2 \succeq r_1$. The symbol $\succeq$ denotes the order among different ranks. Given a training set $\mathcal{D} = \{(\bm{x}_i, y_i)\}_{i=1}^N$, the ordinal regression is to learn a mapping from images to ranks, \emph{i.e.}, $h(\cdot):\mathcal{X} \to \mathcal{Y}$.

Inspired by two ranking-based methods~\cite{chen2017using,niu2016ordinal}, we decompose ordinal regression into a series of binary classifications. Specifically, the ordinal regression with $K$ ranks is decomposed into $K-1$ binary classifiers $\{f_k\}_{k=1}^{K-1}$. For each $r_k\in \{r_1, r_2, \dots, r_{K-1}\}$, a binary classifier is constructed to predict whether the rank of a sample $y_i$ is greater than that of $r_k$. The final rank of an unknown test sample is determined by summarizing all results of the $K-1$ binary classifiers.

To train the $k$-th binary classifier $f_k$, the given dataset $\mathcal{D}$ is divided into two subsets (one positive class and one negative class) determined by whether the age is greater than $k$, \emph{i.e.},
\begin{equation}
D_k^+ = \Big\{(\bm{x}_i, 1)\Big|y_i > r_k\Big\}, D_k^- = \Big\{(\bm{x}_i, 0)\Big|y_i \le r_k\Big\}.
\end{equation}
All $K-1$ binary classifiers are well-trained with their respective training datasets. The age of the test sample $\bm{x_i}$ is predicted as follows:
\begin{equation}
h(\bm{x_i}) = r_1 + \eta\sum_{k=1}^{K-1}\mathbf{1}\Big({f_k(\bm{x_i})} > 0.5\Big),
\end{equation}
where $f_k(\bm{x_i})\in [0, 1]$ is the output probability of the $k$-th classifier for the sample $\bm{x_i}$ (\emph{i.e.}, the $k$-th output of GL-CNN); $\eta$ is the partitioning interval; and $\mathbf{1}(\cdot)$ denotes the truth-test operator, which is 1 if the inner condition holds and 0 otherwise.

\subsection{GL-CNN}
Fig.~\ref{structure} presents an overview of the proposed GL-CNN for gait-based age estimation. The proposed network comprises a single global and three local CNNs, followed by three fully connected layers with $K-1$ outputs. Next, we describe the network in detail.

Grayscale GEIs of size $128\times 88$ are input to the global network. Considering that different parts of a gait take on different local behaviors, we crop the GEI template into three parts: head, body, and feet. In the OULP-Age dataset~\cite{xu2017isir}, the gait images of various people are detected, cropped, aligned, and resized to a uniform silhouette template of same height. In this study, the three parts are cropped using the following three boxes without overlap: $22\times 88$, $48\times 88$, and $58\times 88$. Then, three local networks are designed to learn finer details from these three parts separately. More specifically, there are three convolutional layers in both global and local sub-networks. At the first convolutional layer, 32 filters of size $7\times 7$ with a stride of 1 are applied to the input images, followed by a Leaky Rectified Linear Unit (LeakyReLU)~\cite{maas2013rectifier}. Then, a max-pooling operation with filters of size $2\times 2$ applied with a stride of 2 is used to emphasize the strongest responsive points in the feature maps. Similar operations are performed at the second and third convolutional layers with different filter sizes (Fig.~\ref{structure}). Note that we concatenate the three local feature maps from the second convolution layers along the height dimension to form new local feature maps in a local network. The local network is further concatenated with the feature maps from the global network along the channel dimension.

Then, there are three fully connected layers, as shown in Fig.~\ref{structure}. Among them, F4 is the first fully connected layer in which the feature maps are flattened into a feature vector. There are 1024 neurons in F4, followed by a LeakyReLU and a dropout layer~\cite{srivastava2014dropout}. F5 is the second fully connected layer with 1024 neurons. The second fully connected layer receives the output from F4, followed by a LeakyReLU and another dropout layer. F6 is the third fully connected layer with $K-1$ neurons. This layer receives the output from F5, followed by a LeakyReLU and a dropout layer. Through a sigmoid layer, the $K-1$ output corresponds to the predictive probabilities from $K-1$ binary classifiers. Typically, the network parameters are optimized by minimizing the objective function.

\subsection{ODL}
Here we cast the age label $y_i$ as $\bm{t}_i = [t_i^1, t_i^2, \dots, t_i^{K-1}]^T$ for $K-1$ binary classifiers, where  $t_i^k=\mathbf{1}(y_i > r_k)$. We employ cross-entropy loss as the loss function for these binary classifiers. The loss can be calculated as
\begin{small}
\begin{equation}
\mathcal{L}_c = - \frac{1}{N}\sum_{i=1}^{N}\sum_{k=1}^{K-1}\Big(t_i^k\log(o_i^k) + (1 - t_i^k)\log(1-o_i^k)\Big),
\end{equation}
\end{small}
where $o_i^k$ is the output value of the $k$-th binary classifier for the $i$-th sample. However, the cross-entropy loss optimizes these binary classifiers separately, resulting in a discrepancy between different binary classifications, as shown in Fig.~\ref{toy_example}.

To fully utilize the inner relationships among these $K-1$ outputs, we consider these outputs as a probability distribution and then propose a distribution loss, \emph{e.g.}, $EMD^2$~\cite{hou2016squared}, to penalize the discrepancy between the output distribution and ground-truth distribution. First, the output values are transformed into the probability value by the softmax function,
\begin{equation}
\bar{p}_i^k = \frac{\exp(o_i^k)}{\sum_{j=1}^{K-1}\exp(o_i^j)}, \ \ p_i^k = \frac{\exp(t_i^k)}{\sum_{j=1}^{K-1}\exp(t_i^j)}.
\end{equation}
Then, the $EMD^2$ loss is defined as
\begin{equation}
\mathcal{L}_{\mathrm{EMD}} = \frac{1}{N}\sum_{i=1}^{N}\sum_{k=1}^{K-1}\Big\{CDF_k(\bm{\bar{p}}_i) - CDF_k(\bm{p}_i)\Big\}^2,
\end{equation}
where $\bm{\bar{p}_i} = [\bar{p}_i^1, \bar{p}_i^2, \cdots, \bar{p}_i^{K-1}]^T$ and $\bm{p}_i = [p_i^1, p_i^2, \cdots, p_i^{K-1}]^T$ are the probability distributions corresponding to the $i$-th output $\bm{o}_i$ and the $i$-th ground truth $\bm{t}_i$, respectively. $CDF(\cdot)$ is a cumulative density function of its input, and $CDF_k(\cdot)$ is the $k$-th element of the CDF of its input.

Finally, we propose an ODL by combining the cross-entropy loss with the $EMD^2$ loss. This loss function is easily embedded into the GL-CNN architecture for end-to-end learning. The ODL is
\begin{equation}\label{total_loss}
\mathcal{L}  = \mathcal{L}_c + \lambda\mathcal{L}_{\mathrm{EMD}}
\end{equation}
where $\lambda$ is a hyper-parameter that controls the influence of $\mathcal{L}_{\mathrm{EMD}}$ in the joint loss.

\subsection{Learning ODR-GLCNN}
One advantage of using Eq.~\eqref{total_loss} is that the ODL can simultaneously learn each binary classification and the inner relationship between these binary classifications. For the $i$-th sample $\bm{x}_i$, the gradient of our loss can be derived as
\begin{equation}
   \frac{\partial \mathcal{L}}{\partial \bm{W}}=\frac{\partial \mathcal{L}_c}{\partial \bm{o}_i}\frac{\partial \bm{o}_i}{\partial \bm{W}} + \lambda\frac{\partial \mathcal{L}_{\mathrm{EMD}}}{\partial \bm{\bar{p}}_i}\frac{\partial \bm{\bar{p}}_i}{\partial \bm{W}},
\end{equation}
where $\bm{W}$ represents the network parameters, and $\frac{\partial \bm{o}_i}{\partial \bm{W}}$ and $\frac{\partial \bm{\bar{p}}_i}{\partial \bm{W}}$ can be derived through the standard backpropagation method. For the $k$-th element of $\bm{o}_i$, the gradient can be derived as
\begin{equation}
\label{cross_loss_derived}
  \frac{\partial \mathcal{L}_c}{\partial o_i^k}= - \Big(\frac{t_i^k}{o_i^k}-\frac{1-t_i^k}{1-o_i^k}\Big).
\end{equation}
For the $k$-th element of $\bm{\bar{p}_i}$, the gradient can be derived as
\begin{equation}
\label{emd_loss_derived}
\begin{aligned}
\frac{\partial \mathcal{L}_{\mathrm{EMD}}}{\partial \bar{p}_i^k}
&=\frac{\partial \sum_{j=1}^{K-1}(CDF_j(\bar{p}_i)-CDF_j(p_i))^2}{\partial \bar{p}_i^k} \\
&=\frac{\partial \sum_{j=1}^{K-1}(\sum_{\ell=1}^{j}(\bar{p}_i^\ell-{p}_i^\ell))^2}{\partial \bar{p}_i^k} \\
&=2\sum_{j=k}^{K-1}\sum_{\ell=1}^j(\bar{p}_i^\ell -{p}_i^\ell ),
\end{aligned}
\end{equation}
where $k\in\{1, 2, \dots, K-1\}$. Eq.~\eqref{cross_loss_derived} indicates that the gradient of the cross-entropy loss is only related to the output value of each binary classification and its corresponding ground truth, ignoring the intrinsic correlation for their binary classifiers. In contrast, the output value of each classification is considered when computing the gradient of a specific binary classification in $EMD^2$ loss, as shown in Eq.~\eqref{emd_loss_derived}. Therefore, the ODL considers each binary classification and can utilize the inner relationships among them.

\section{Experiments}
In this section, we describe the experimental setting and demonstrate the effectiveness of the proposed method by comparing it with state-of-the-art methods and performing a set of ablative studies on the OULP-Age gait dataset~\cite{xu2017isir}. In addition, we evaluate the generalization ability of the proposed approach to other tasks, \emph{e.g.,} facial age estimation on the MORPH Album II dataset~\cite{ricanek2006morph}.

\subsection{Experimental setting}
\subsubsection{Data preparation}
\begin{figure}
\centering
\includegraphics[width=1\linewidth,height=0.58\linewidth, clip=true, trim=20 0 20 10]{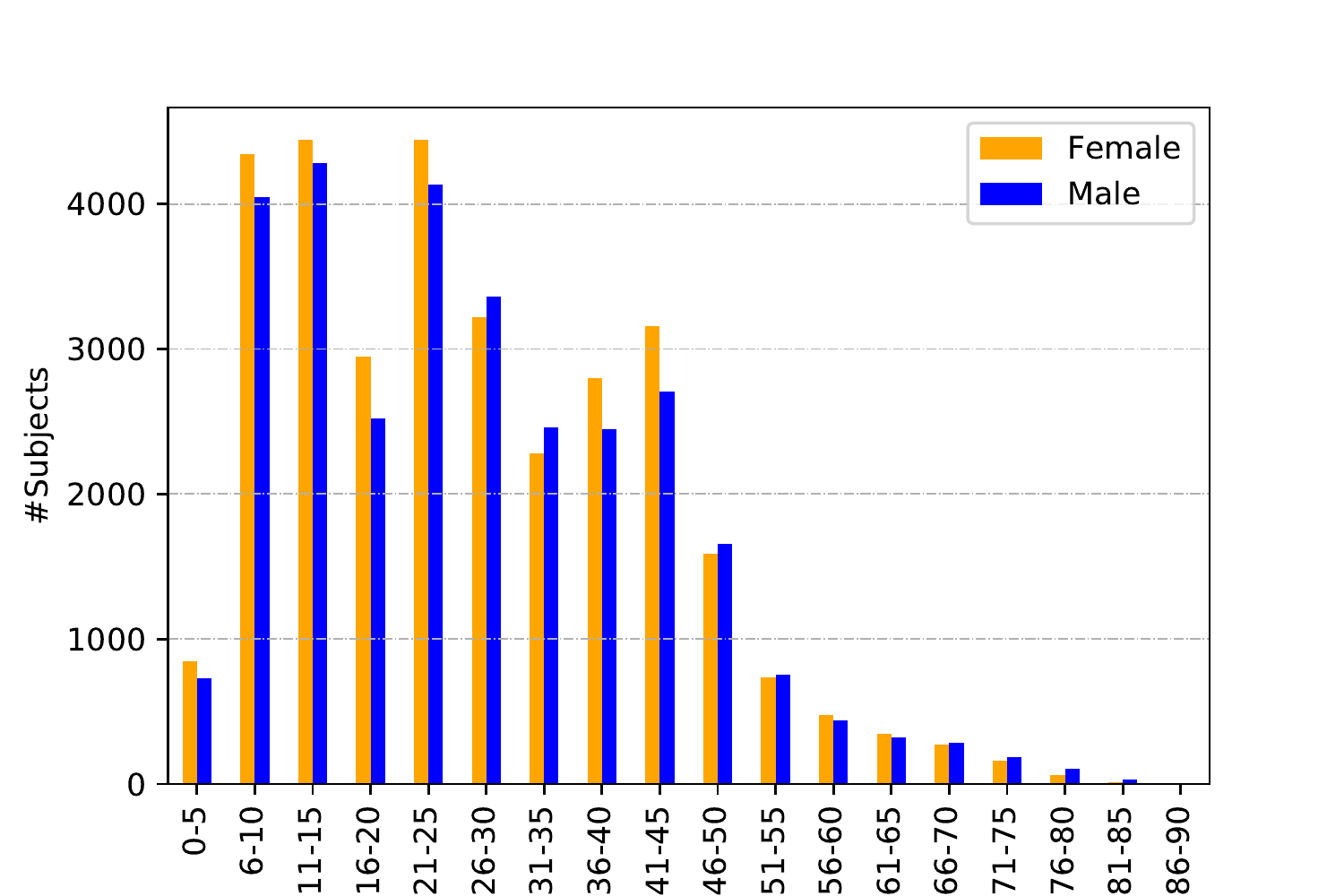}
\caption{Age and gender distribution for the OULP-Age dataset. There are 63,846 well-labeled GEIs (31,093 males and 32,753 females, age 2 to 90 years).}
\label{data_statistics}
\end{figure}

\textbf{OULP-Age} is the largest gait dataset in the world to date, containing 63,846 GEIs (31,093 males and 32,753 females; age 2 to 90 years; sample size $128 \times 88$). Gender distribution by age in five-year intervals is shown in Fig.~\ref{data_statistics}. As Xu \emph{et at.}~\cite{xu2017isir} suggests, the OULP-Age dataset is averagely divided into two disjoint subsets (training set and test set). The training set contains 15,596 males and 16,327 females, and the test set contains 15,497 males and 16,426 females. Note that both datasets have similar age distribution.

Note that \textbf{USF}~\cite{sarkar2005humanid} is another popular gait dataset for age estimation. However, it is too small (only 122 subjects) and rough (age labels are not very accurate) to be used for evaluating the performance of deep-learning-based gait age estimation. Therefore, we trained the proposed method on the OULP-Age dataset and used the USF dataset only for testing.

\textbf{MORPH Album II} is one of the largest and most popular longitudinal face databases for age estimation in the public domain (55,134 face images, 13,617 subjects, age 16 to 77 years) ~\cite{ricanek2006morph}. To further demonstrate the effectiveness of the ODL, we evaluated the proposed loss on the MORPH Album II dataset. Following the previous studies ~\cite{niu2016ordinal,chen2017using,pan2018mean,shen2018deep}, we used the five-fold random split (RS) protocol to evaluate the facial age estimation performance. All face images were aligned based on five facial landmarks detected using the open-source SeetaFaceEngine face alignment algorithm\footnote{https://github.com/seetaface/SeetaFaceEngine} and then were resized to $256\times 256\times 3$.

\subsubsection{Evaluation metrics}
The age estimation performance was evaluated using the mean absolute error (MAE) and the cumulative score (CS). MAE represents the average of the absolute errors between the predicted age and the ground truth over all test samples. MAE is defined as $\frac{1}{N}\sum_{i=1}^{N}\Big|h(\bm{x}_i) - y_i\Big|$, where $N$ is the total number of test samples. CS is calculated as $CS(k) = \frac{N_k}{N} \times 100\%$, where $N_k$ is the number of test samples whose absolute error between the estimated age and the ground truth is not greater than $k$ years. CS reveals a consistent performance by computing the accuracy of the evaluated model at different $k$-levels.

\subsection{Gait-based age estimation results}
\subsubsection{Implementation details}
In our experiments, we utilized our GL-CNN, a CNN (comprising a global part, as shown in Fig.~\ref{structure}), and a VGG16~\cite{parkhi2015deep,simonyan2014very} as three backbone networks. We used Adam~\cite{kinga2015method} with a learning rate of $0.0001$, beta1 0.5, beta2 0.999, weight decay of $0.00001$, batch size of 300, and maximal epochs of 300 for CNN and GL-CNN. Following a previous study ~\cite{pan2018mean}, we used stochastic gradient descent with a learning rate of $0.0001$, weight decay of $0.0001$, batch size of 300, and maximal epochs of 100 for VGG16. The learning rate was reduced by multiplying the effective rate by 0.1 every 15 epochs. To make the grayscale GEIs suitable for VGG16, we copied the GEIs three times as RGB channels to feed into the VGG16, which was pre-trained on ImageNet~\cite{russakovsky2015imagenet}. In addition, the weight coefficient of the $EMD^2$ loss term in Eq.~\eqref{total_loss} was set to $\lambda = 10$, which was tuned according to the model performance. All experiments were implemented on PyTorch with four GeForce GTX 1080 Ti GPUs.

\subsubsection{Comparison of the proposed method to state-of-the-art methods on the OULP-Age dataset}
We compared the proposed method with state-of-the-art methods, including classification-based methods (e.g., MLG~\cite{lu2010gait}), regression-based methods (e.g., GPR~\cite{makihara2011gait}, SVR~\cite{smola2004tutorial}, and ASSOLPP~\cite{li2018gait}), and age manifold learning-based methods (e.g., OPLDA and OPMFA~\cite{lu2010ordinary}). In addition, we implemented a deep learning method as a baseline, i.e., \textbf{VGG16 + Mean-Variance}, to validate the effectiveness of the proposed method. This method ~\cite{pan2018mean} achieves an outstanding performance in the field of face-based age estimation.

\begin{table}[h]
\small
  \begin{center}
    \begin{tabular}{|l|c|c|}
    \hline
      Methods                               & MAE         & CS ($k=5$) \\ \hline
      SVR~\cite{smola2004tutorial}         & 7.66        & 41.40\%          \\ \hline
      MLG~\cite{lu2010gait}                & 10.98       & 43.40\%          \\ \hline
      OPLDA~\cite{lu2010ordinary}          & 8.45        & 36.50\%          \\ \hline
      OPMFA~\cite{lu2010ordinary}          & 9.08        & 34.70\%          \\ \hline
      GPR~\cite{makihara2011gait}          & 7.30        & 43.60\%          \\ \hline
      ASSOLPP~\cite{li2018gait}            & 6.78        & 53.00\%          \\ \hline
      VGG16 + Mean-Variance~\cite{pan2018mean} & 5.59        & 60.46\%          \\ \hline
      \textbf{ODR-GLCNN (Ours)}            & {\bf 5.12}  & \bf{66.95\%}     \\ \hline
    \end{tabular}
  \end{center}
  \caption{Comparison of the age estimation MAEs with the proposed approach and state-of-the-art methods on the OULP-Age dataset.}\label{compare_prior_MAE}
\end{table}

\begin{figure}[h]
\centering
\includegraphics[width=1\linewidth,height=0.65\linewidth, clip=true, trim=40 32 80 40]{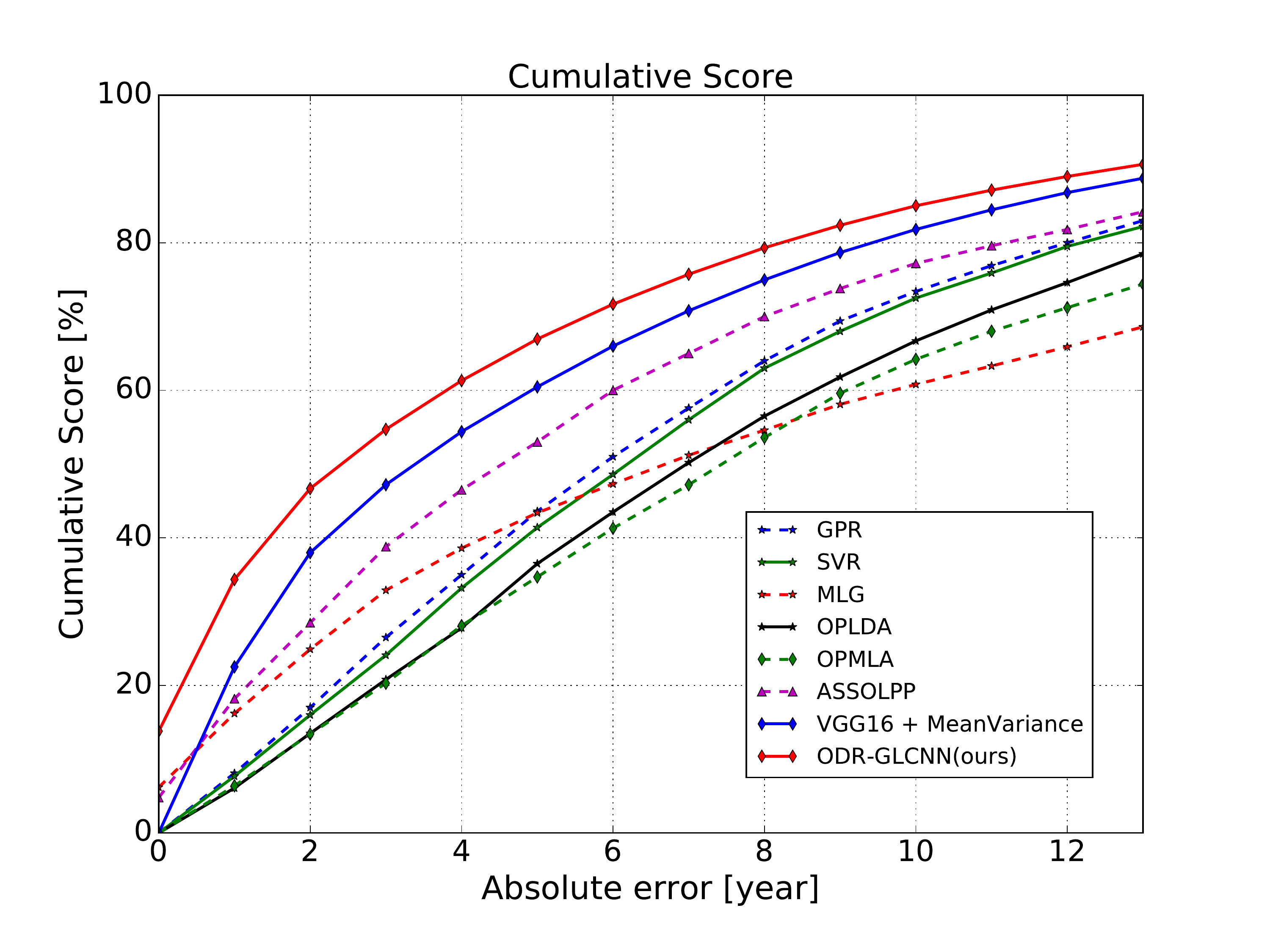}
\caption{Comparisons of age estimation CSs by the proposed approach and state-of-the-art methods on the OULP-Age dataset.}
\label{compare_prior_cs_result}
\end{figure}

Table~\ref{compare_prior_MAE} shows the result obtained by eight methods on the OULP-Age dataset. This suggests that CNN-based methods, such as~\cite{pan2018mean}, perform better than traditional methods in terms of MAE~\cite{li2018gait,lu2010gait,lu2010ordinary,makihara2011gait,smola2004tutorial}. CNN-based methods demonstrate better performance because they have many more parameters and learn more representative features with end-to-end training. The proposed method performs the best among all approaches because it benefits from both a more representative feature extraction network (GL-CNN) and a unique loss function (the ODL), which can learn each binary classification of ordinal regression and the inner relationships among them. In addition, as shown in Fig.~\ref{compare_prior_cs_result}, the CS results on the OULP-Age dataset further demonstrate that the proposed approach performs consistently better than the other state-of-the-art methods.

\begin{figure*}[h]
\centering
\includegraphics[width=0.98\linewidth,height=0.6\linewidth, clip=true, trim=80 10 100 10]{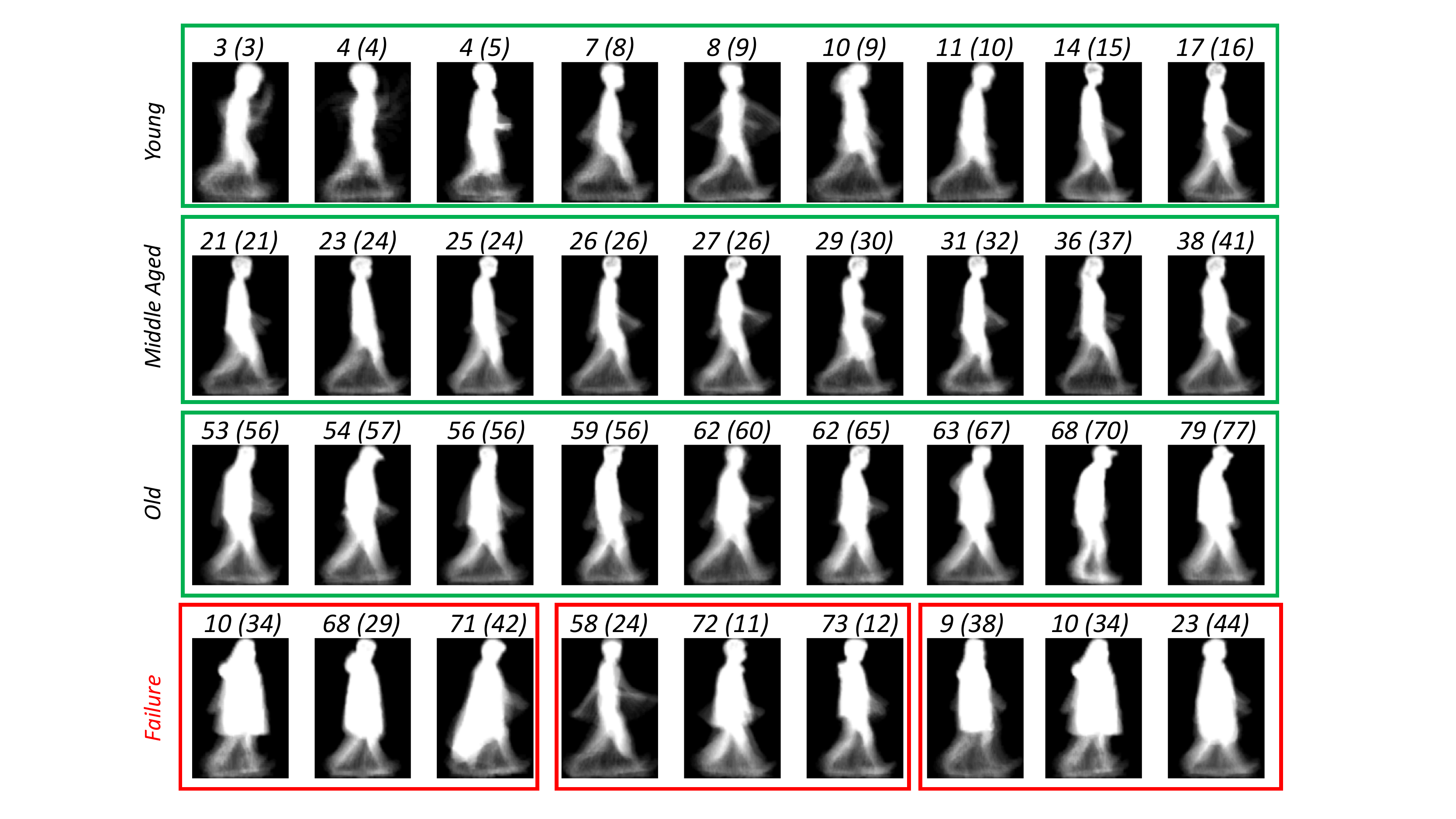}
\caption{Examples of gait-based age estimation results by using the proposed approach on the OULP-Age dataset. The top three rows show nine successful age estimation examples (MAE smaller than 5 years) for young, middle-aged, and old subjects, respectively. The last row shows nine failure cases (MAE larger than 20 years). The numbers above each image show the subject’s ground-truth age and estimated age, respectively.}
\label{predict_examples}
\end{figure*}

Some age estimation examples are shown in Fig.~\ref{predict_examples}. As seen, the proposed approach is quite robust for young, middle-aged, and old subjects. However, as is noticeable from the last row of Fig.~\ref{predict_examples}, age estimation accuracy may degenerate when a person wears heavy clothes or when a person is extremely thin or overweight.

\subsubsection{Testing with state-of-the-art methods on the USF dataset}
To further demonstrate the effectiveness of the proposed method, we evaluated the performance of our model on the USF dataset and compared it with state-of-the-art methods. As described above, the USF dataset is too small to train a deep network model. Therefore, we trained our model and other methods on the OULP-Age dataset, and then tested using the USF dataset.
\begin{table}[h]
\scriptsize
  \begin{center}
    \begin{tabular}{|l|c|c|c|c|}
    \hline
    \multirow{2}{*}{Methods} &\multicolumn{2}{c|}{Gallery} &\multicolumn{2}{c|}{Probe A} \\ \cline{2-5}
          & MAE & CS($k=5$) & MAE & CS($k=5$) \\ \hline
      SVR~\cite{smola2004tutorial}         & 8.21 & 37.50\%  & 7.83 & 41.70\%  \\ \hline
      MLG~\cite{lu2010gait}                & 9.45 & 32.80\% & 9.06  & 34.40\%  \\ \hline
      OPLDA~\cite{lu2010ordinary}          & 7.05 & 43.70\% & 6.76  & 51.20\%  \\ \hline
      OPMFA~\cite{lu2010ordinary}          & 6.95 & 47.30\%  & 6.62  & 52.00\%  \\ \hline
      ASSOLPP~\cite{li2018gait}            & 6.81 & 50.50\%  & 6.48   &  52.70\%  \\ \hline
      VGG16 + Mean-Variance~\cite{pan2018mean}& 6.10   & 54.20\%  & 5.93  & 56.30\% \\ \hline
      \textbf{ODR-GLCNN (Ours)}     &   \bf{5.91} & \bf{56.60\%}  & \bf{5.75}  & \bf{59.40\%}\\ \hline
    \end{tabular}
  \end{center}
    \caption{MAE results on the two subsets of the USF dataset using different methods.}\label{test_on_usf}
\end{table}

Table~\ref{test_on_usf} shows that for the USF gallery (probe A) subset, the proposed method outperforms SVR, MLG, OPLDA, OPMFA, ASSOLPP, and VGG16+Mean-Variance with MAE gains of 2.3 (2.08), 3.54 (3.31), 1.14 (1.01), 1.04 (0.87), 0.9 (0.73), and 0.19 (0.18) years old, respectively. The results further indicate that the proposed method consistently outperforms other methods for gait-based age estimation.

\subsubsection{Analyzing the performance of GL-CNN}
We evaluated the performance of the proposed GL-CNN by comparing it with a simple CNN comprising a global part and the VGG16 network, which is widely used for age estimation based on gait. All three networks used cross-entropy loss as the loss function. The MAE and CS ($k=5$) results for these three different networks are shown in Table~\ref{compare_network}.

\begin{table}[h]
\small
  \begin{center}
    \begin{tabular}{|l|c|c|c|}
    \hline
      Network     &  MAE       & CS ($k=5$)   &  Time (ms)     \\ \hline
      CNN         &  5.45      & 64.64\%      &  \bf{7.27} $\times 10^{-2}$ \\ \hline
      VGG16       &  5.63      & 63.92\%      &  21.9 $\times 10^{-2}$      \\ \hline
      GL-CNN (\textbf{Ours})     &  \bf{5.24} & \bf{65.96\%} &  8.99 $\times 10^{-2}$      \\ \hline
    \end{tabular}
  \end{center}
   \caption{Comparisons among different CNN-based methods on the OULP-Age dataset. The test time for a single sample is shown in the last column.}\label{compare_network}
\end{table}

Compared to a simple CNN, GL-CNN achieves better age estimation performance.
Table~\ref{compare_network} shows that 1) compared to CNN and VGG16, GL-CNN achieves the best performance in two criteria; 2) although VGG16 has more parameters, GL-CNN effectively learns more detailed information by combining a global and three local structures, resulting in an improved performance; and 3) the computational cost of GL-CNN is only slightly higher than that of CNN and much smaller than that of VGG16.

\begin{figure}[!t]
\centering
\includegraphics[width=1\linewidth,height=0.5\linewidth, clip=true, trim=20 140 20 123]{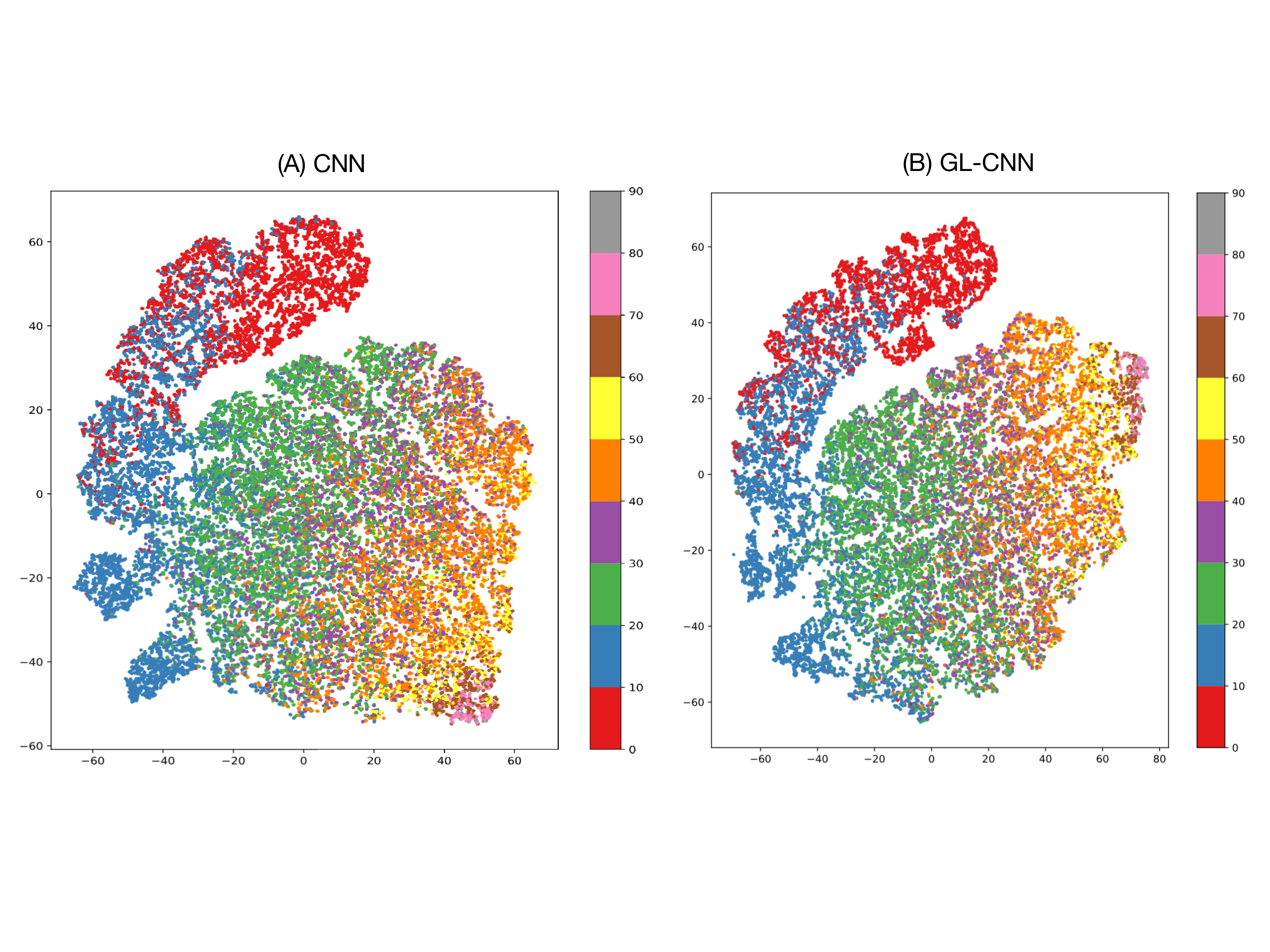}
\caption{Feature visualization of CNN (A) and GL-CNN (B). Network features are reduced from 1024 dimensions to 2 dimensions by a $t$-SNE technique. Ages are divided into nine groups. Different colors represent different age groups.}
\label{visualization}
\end{figure}

To better demonstrate the effectiveness of the proposed network, we visualize CNN and GL-CNN features through $t$-distributed stochastic neighbor embedding~\cite{maaten2008visualizing} ($t$-SNE) technique with perplexity 30, as shown in Fig.~\ref{visualization}. For better visualization, the age label is divided into 9 age groups: \emph{e.g.} $[0\sim10], [11\sim20], [21\sim30], \cdots, [81\sim90]$. As can be seen, both GL-CNN and CNN features appear to maintain a manifold-like structure because the order of ages varies smoothly from left to right. However, after zooming in Fig.~\ref{visualization}, it is evident that the inner age group samples of GL-CNN are denser than those of CNN, particularly in age groups $[21\sim30]$ and $[31\sim40]$, which shows that GL-CNN can achieve fewer age estimation errors because it learns a better feature representation from both global and local structures.

\subsubsection{Comparison of different losses}
To validate the effectiveness of the proposed ODL, we compared it with three widely used losses in gait-based age estimation tasks, \emph{e.g.}, Euclidean, MAE, and cross-entropy losses, by performing age estimation based on the proposed GL-CNN. The MAE and CS ($k=5$) of these losses are reported in Table~\ref{compare_loss}.
\begin{table}[h]
\small
  \begin{center}
    \begin{tabular}{|l|c|c|}
    \hline
      Loss                    &  MAE        & CS ($k=5$)                  \\ \hline
      Euclidean               & 6.73        & 52.95\%                     \\ \hline
      MAE                     & 6.65        & 55.16\%                     \\ \hline
      Squared EMD\cite{hou2016squared}      & 6.39  & 58.34\%      \\ \hline
      Cross-Entropy           & 5.24        & 65.96\%                     \\ \hline
      ODL (\textbf{Ours})  & \bf{5.12}    & \bf{66.95\%} \\ \hline
    \end{tabular}
  \end{center}
  \caption{Comparison of different losses with the proposed GL-CNN on the OULP-Age dataset.}\label{compare_loss}
\end{table}

It can be seen that cross-entropy loss outperforms Euclidean loss and MAE loss for the age estimation tasks. Euclidean and MAE losses easily lead to over-fitting and do not consider the ordinal information between age labels. In contrast, the proposed ODL incorporates the inner relationship between the binary classifications by using a distribution loss, i.e., $EMD^2$ loss, resulting in better predictive performance.

In addition, to demonstrate the difference between our proposed ODL and the squared EMD loss in ~\cite{hou2016squared}, we compare them on the OULP-Age dataset. Following~\cite{hou2016squared}, the squared EMD loss is utilized on a classification task in these experiments. The comparison results, shown in Table~\ref{compare_loss}, indicate that the combination of squared EMD loss and ordinal regression is more suitable for the gait-based age estimation task. The results indicate that utilizing squared EMD loss with ordinal regression is reasonable and effective.

\subsubsection{Influence of gender}
\begin{figure}[!ht]
\centering
\includegraphics[width=1\linewidth,height=0.44\linewidth, clip=true, trim=20 350 250 40]{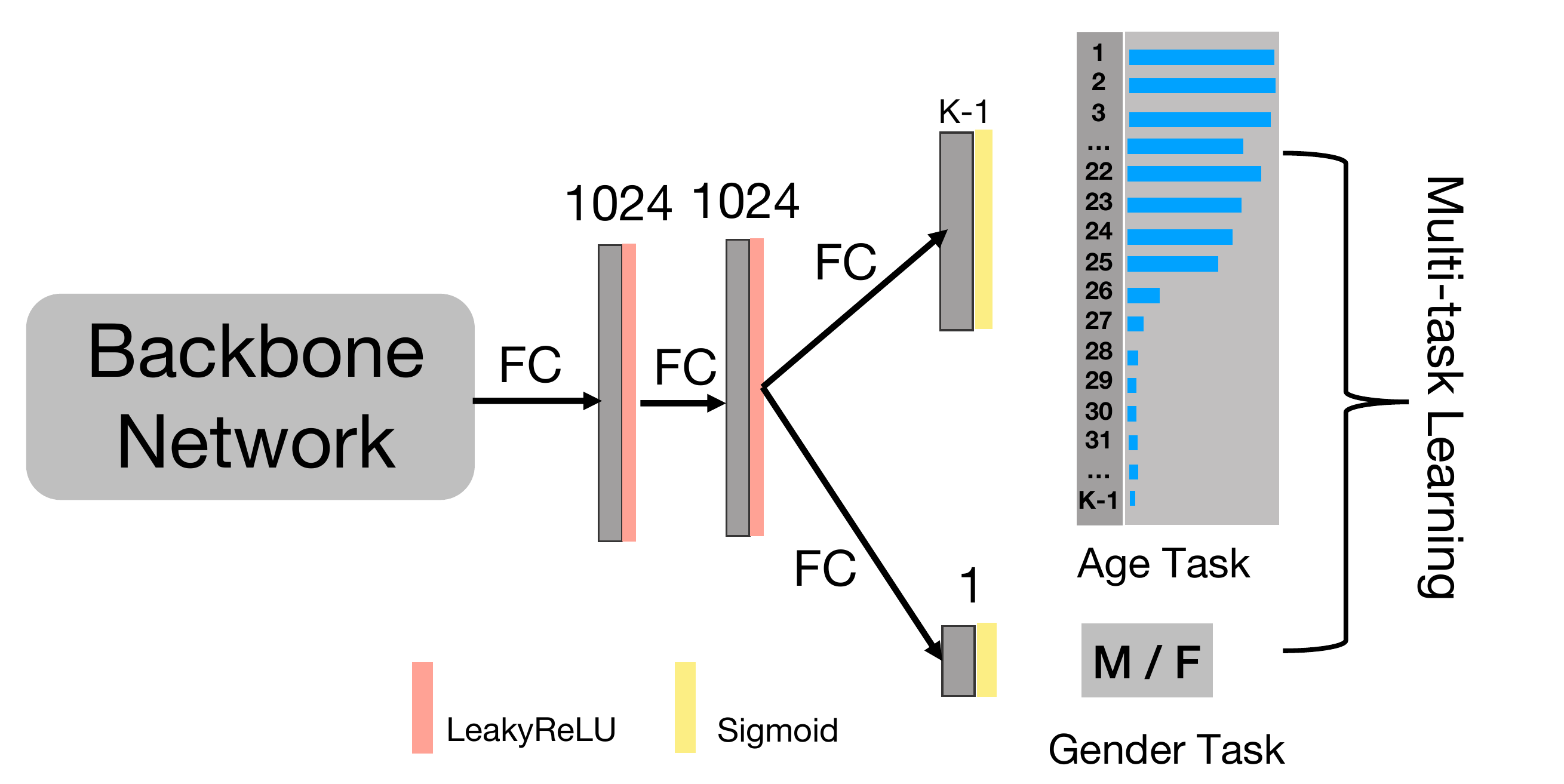}
\caption{Multi-task structure for age and gender estimation tasks.}
\label{multi-task_structure}
\end{figure}

We realize that gender is correlated with age because human gait varies between males and females even within the same age group, as shown in Fig.~\ref{gait_examples}. To better utilize the relationship between age and gender, we embedded a multi-task technique into the CNN-based framework. As shown in Fig.~\ref{multi-task_structure}, specifically, we investigated a gender classification task for the proposed method and three other CNN-based methods. As a binary classification, the gender loss is defined as
\begin{small}
\begin{equation}
\mathcal{L}_{gender} = - \frac{1}{N}\sum_{i=1}^{N}\Big(g_i\log(\hat{g_i}) + (1-g_i)\log(1-\hat{g_i})\Big),
\end{equation}
\end{small}
where $g_i$ is the ground truth of gender for the $i$-th sample and $\hat{g_i}$ is the corresponding predicted value.

\begin{table}[h]
\small
  \begin{center}
    \begin{tabular}{|l|c|c|c|}
    \hline
      Method                    &  w/o gender    & w/ gender       & acc.  \\ \hline
      CNN + Euclidean           &  6.96          &  \bf{6.82}      & 96.70\%   \\ \hline
      CNN + Cross-Entropy       &  5.40          &  \bf{5.34}      & 97.20\%   \\ \hline
      VGG16 + Mean-Variance     &  5.59          &  \bf{5.52}      & 96.70\%   \\ \hline
      ODR-GLCNN (\textbf{Ours}) &  5.12          &  \bf{5.06}      & \bf{97.80\%}   \\ \hline
    \end{tabular}
  \end{center}
  \caption{Influence of human gender for gait-based age estimation in terms of age MAE and gender accuracy.}\label{compare_gender}
\end{table}

Table~\ref{compare_gender} indicates that gender information indeed improves the performance of gait-based age estimation. Moreover, the accuracy of gender classification with the proposed method is 97.8\%, implying that, as a byproduct, the proposed network can accurately predict gender from the gait.

\subsection{Facial age estimation}
We applied the proposed ODL to facial age estimation using the MORPH Album II dataset, and compared the results with those of state-of-the-art methods~\cite{niu2016ordinal,rothe2018deep,chen2017using,shen2017label,pan2018mean,shen2018deep,tan2017efficient,li2019bridgenet}. Following  previous studies, ~\cite{pan2018mean,shen2018deep}, we also utilized VGG16, pre-trained with ImageNet~\cite{russakovsky2015imagenet}, as the backbone network with the proposed ODL. The results of individual approaches in terms of MAE and CS are reported in Table~\ref{compare_face}. As can be seen, our approach achieves better prediction performance than the state-of-the-art method DRF~\cite{shen2018deep}, which suggests that the proposed approach can be well-generalized to the facial age estimation task. In addition, the results obtained using ODL ($\lambda=0.25$) are better than those obtained using a single cross-entropy loss ($\lambda=0$ when calculating ODL), which indicates that ODL is more effective in learning the ordinal relationships among different ages than a single cross-entropy loss.
\begin{table}[h]
\footnotesize
  \begin{center}
    \begin{tabular}{|l|c|c|c|}
    \hline
      Method                                   & MAE       & CS ($k=5$)    & Protocol \\ \hline
      OR-CNN~\cite{niu2016ordinal}              & 3.27      & 73.0\%*       & RS     \\ \hline
      DEX~\cite{rothe2018deep}                  & 3.25      & --           & RS     \\ \hline
      Ranking-CNN~\cite{chen2017using}          & 2.96      & 85.0\%*       & RS     \\ \hline
      VGG16 + Mean-Variance~\cite{pan2018mean}  & 2.41      & 90.0\%*       & RS     \\ \hline
      AGEn~\cite{tan2017efficient}              & 2.93      & --           & RS     \\ \hline
      dLDLF~\cite{shen2017label}                & 2.24      & --           & RS     \\ \hline
      DRFs~\cite{shen2018deep}                  & 2.17      & 91.3\%        & RS     \\ \hline
      BridgeNet~\cite{li2019bridgenet}          & 2.38      & --        & RS     \\ \hline
      VGG16 + ODL($\lambda=0$)                  & 2.30      & 91.1\%        & RS     \\ \hline
      VGG16 + ODL($\lambda=0.25$)(\textbf{Ours})& \bf{2.16} & \bf{92.9\%}   & RS     \\ \hline
    \end{tabular}
  \end{center}
  \caption{Comparison between proposed approach and state-of-the-art methods on the MORPH Album II dataset in terms of MAE and CS values (*: the value is read from the reported CS curve). RS represents the five-fold {\bf R}andom {\bf S}plit protocol.}\label{compare_face}
\end{table}

\section{Conclusion}
In this paper, we proposed an ordinal distribution regression, including a novel network GL-CNN and a useful loss function ODL, for gait-based age estimation. Specifically, the GL-CNN consisting of one global and three local sub-networks was constructed to extract more representative gait features. And the proposed ODL incorporating cross-entropy loss and $EMD^2$ loss was found to be more effective in learning the ordinal relationships among different ages than a single cross-entropy loss. We also found that if the gender information is available for training, embedding a multi-task strategy into the proposed framework can more or less improve the age estimation performance. Experiments on the OULP-Age and MORPH Album II datasets show that the proposed approach outperforms state-of-the-art methods on gait-based age estimation and generalize well for facial age estimation tasks.

In the future, it is worth studying how to utilize temporal information or cross-view information~\cite{he2019multi,Takemura2018,wu2017comprehensive} of gait sequences to improve the accuracy and effectiveness of gait-based age estimation.

\section*{Acknowledgements}This work supported in part by the National Natural Science Foundation of China (NSFC61673118), Shanghai Municipal Science and Technology Major Project (No.2018SHZDZX01), ZJLab, and Shanghai Pujiang Program (No.16PJD009). We are grateful to the reviewers and the Associate Editor for their constructive comments.

\small
\bibliographystyle{ieee}

\end{document}